\pdfoutput=1

\documentclass[11pt]{article}

\usepackage{ACL2023}

\usepackage{setspace}

\usepackage{times}
\usepackage{latexsym}

\usepackage[T1]{fontenc}

\usepackage[utf8]{inputenc}

\usepackage{microtype}

\usepackage{inconsolata}

\usepackage{booktabs}
\usepackage{csquotes}
\usepackage{multirow}

\usepackage{listings}

\title{Integrating Expert Labels into LLM-based Emission Goal Detection:\\ Example Selection vs 
Automatic Prompt Design}

\author{Marco Wrzalik$^1$, Adrian Ulges$^1$, Anne Uersfeld$^1$, Florian Faust$^2$, Viola Campos$^1$ \\
  $^1$ RheinMain University of Applied Sciences (Wiesbaden, Germany) \\ 
  $^2$ Sociovestix Labs Ltd. (Kaiserslautern, Germany) \\ $^1$\texttt{firstname.lastname@hs-rm.de}, 
  $^2$ \texttt{firstname.lastname@sociovestix.com}}
  
\begin{document}

\maketitle

\begin{abstract}
We address the detection of emission reduction goals in corporate reports, an important task for monitoring companies' progress in addressing climate change.
Specifically, we focus on the issue of integrating expert feedback in the form of labeled example passages into LLM-based pipelines, and compare the two strategies of (1) a dynamic selection of few-shot examples and (2) the automatic optimization of the prompt by the LLM itself. 

Our findings on a public dataset of 769 climate-related passages from real-world business reports indicate that automatic prompt optimization is the superior approach, while combining both methods provides only limited benefit. Qualitative results indicate that optimized prompts do indeed capture many intricacies of the targeted emission goal extraction task.
\end{abstract}

%

\section{Introduction}
\label{sec:intro}
The urgency of the climate crisis necessitates immediate action across various sectors of the global economy. These efforts are targeted at {\it Net Zero}, i.e. achieving a balance between emitted and removed greenhouse gases, particularly CO$_2$.
Policies such as the European Union's \textit{Green Taxonomy} aim to redirect financial investment flows toward sustainable businesses, setting incentives for companies committing to concrete emission reduction goals. 

To evaluate and monitor these commitments, analysts must review extensive corporate documentation, including annual reports, sustainability reports, and stewardship disclosures. This manual process of locating and extracting relevant data, often referred to as \textit{carbon accounting}, is labor-intensive and time-consuming. 
Identifying such  climate goals is a surprisingly intricate task, as analysts have to distinguish concrete, binding, self-imposed and quantitative emission goals from vague statements, greenwashing,  externally mandated requirements, goals {\it related to} climate change (such as "moving out of coal"), etc. 
Overall, the task not only requires high precision but also poses unique challenges due to the nuanced language and diverse reporting styles used by companies.


To increase the efficiency of carbon accounting, Large Language Models (LLMs)  have emerged 
as powerful tools. Given a {\it prompt} consisting of task-specific instructions and a  text passage to analyze, the LLM outputs whether the passage contains an emission goal. 
When integrating such LLMs into practical workflows, their outputs are inspected and -- in case of errors --  corrected by analysts, resulting in a set of labeled "challenge samples" that grows over time. These examples offer an interesting option for \textit{in-context learning} (ICL), i.e. to improve the LLM's accuracy by utilizing the examples to improve the prompt, without applying fine-tuning to the model.


In this paper, we investigate ICL strategies to improve LLMs with expert knowledge in form of labeled examples, focusing on the task of emission goal detection. Particularly, we compare two approaches: (1) \textit{Example selection}, which incorporates a limited number of few-shot examples into the prompt. These exemplify the desired behavior, and 
are selected dynamically to resemble the input passage.
(2) 
\textit{Automatic Prompt Design}, in which the LLM adjusts its own instructions.
This approach applies an iterative optimization process in which erroneous challenge cases are inspected, reflected, and new, refined instructions are generated and evaluated.

While both techniques hold promise, their effectiveness in real-world applications has not been compared extensively. 
We hope to fill this gap with the following  contributions:

\begin{enumerate}
    \item We conduct a comprehensive comparison of example selection and automatic prompt design on the task of emission goal classification, using  769 passages from the public \textit{NetZeroFacts} dataset~\cite{wrzalik24}.
    \item We highlight the strengths and limitations of each prompting strategy, guiding practitioners applying LLMs to sustainability classification tasks. Specifically: (1) example selection improves results, especially with weak prompts; (2) auto-prompting yields greater overall gains; and (3) in automatic prompt design, few-shot examples provide limited additional benefits.
\end{enumerate}


\section{Related Work}
This section reviews existing literature on prompting strategies that make use of labeled data, focusing on example selection for few-shot learning and automatic prompt design. 

\subsection{Example Selection for Few-Shot Learning}

\citet{liu22} investigated the selection of good in-context examples for GPT-3 and found that semantically similar examples, chosen based on proximity in embedding space, significantly improve model performance. 
\citet{rubin22} proposed a contrastive learning-based method that learns to retrieve task-specific examples, showing significant performance improvements by optimizing the input-output pairs for correct predictions. 
\citet{su23} introduced a selective annotation framework that enhances few-shot learning by strategically selecting diverse and representative examples from a small annotated pool. Their graph-based approach demonstrates that such careful example selection can lead to significant performance improvements across various natural language processing tasks.
 \citet{zhang22} framed example selection as a reinforcement learning problem, proposing a Q-learning-based approach to actively select examples. This method shifts from similarity-based retrieval to learning a policy that optimizes example selection for few-shot learning, demonstrating moderate improvements on downstream tasks.
In this work, we focus on the selection of semantically similar examples as motivated by \citet{liu22}.

\subsection{Automatic Prompt Design}

The task of optimizing prompts for LLMs has received increasing attention, with several approaches leveraging the model's ability to self-improve: \citet{shin20} introduced \textit{AutoPrompt} uses a 
gradient-guided search to iteratively refine trigger tokens, resulting in competitive performance compared to manually designed prompts for tasks like natural language inference.
\citet{zhou23} proposed \textit{Automatic Prompt Engineer} (APE), 
which iterates over LLM-generated candidate prompts, selecting and refining those that lead to the best performance based on task-specific score functions, often outperforming human-generated prompts. 
Similarly, \citet{pryzant23} introduced \textit{ProTeGi}, a method that optimizes prompts 
based on LLM-generated rasonings over erroneous examples. The performance of these reasonings is estimated and stears a beam search that explores multiple candidate options. 
\citet{yang24} utilize the LLM itself as a general-purpose optimizer capable of refining prompts and solving various tasks. Their method, OPRO, iteratively generates and evaluates prompts based on the history of previous attempts, framing prompt optimization as a general meta-optimization task. 
Finally, \textit{Intent-based Prompt Calibration} (IPC) by \citet{levi24} introduces the auto-generation of synthetic challenge cases to calibrate prompts, which are then labeled by the expert.
Our work explores automatic prompt design -- as outlined above -- from a practitioners perspective. Specifically, we follow a simplified variant of \citet{pryzant23}'s {\it ProTeGi}, which (instead of beam search) performs a greedy search.

\begin{table*}[ht!]
    \centering
    \caption{Results of Few-shot Prompting (\%, including example selection in Row 4 ("Similar")).}
    \label{tab:fewshot}
    \begin{small}
    \begin{tabular}{lcccc|cccc}
    \toprule
    & \multicolumn{4}{c}{simple instructions} & \multicolumn{4}{c}{expert instructions} \\
    \cmidrule(lr){2-5} \cmidrule(lr){6-9}
    \textsc{Examples} & Acc & Prec & Rec & F1 & Acc & Prec & Rec & F1 \\
    \midrule
    Zero-shot & 56.2 & 52.2 & 100.0 & 68.6 & 81.2 & 77.1 & 86.1 & 81.4 \\
    Static  & 72.3 & 63.8 & 97.0 & 77.0 & 82.0 & 90.8 & 69.4 & 78.7 \\
    Random  & 66.2 & 58.9 & 98.6 & 73.7 & 80.5 & 75.7 & 89.5 & 81.5 \\
    Similar & 69.8 & 62.3 & 92.9 & 74.6 & 87.0 & 87.2 & 85.4 & 86.3 \\
    \bottomrule
    \end{tabular}
    \end{small}
\end{table*}






\subsection{Emission Information Extraction}

The detection and extraction of SDG (Sustainable Development Goals) related information has recently gained traction as a research field: 
\citet{spokoyny2023answeringclimatequestionnairesunstructured}
bridge the gap between structured reporting (in form of questionnaires) and unstructured reporting in form of text: Their ClimaBench benchmark challenges NLP models to extract climate-related information from company reports and thus auto-fill questionnaires.
\citet{schimanski2024bridging} pretrain specific BERT models for environmental, social and governance aspects on 13.8 million corporate disclosures and curate three balanced 2k‑document test collections for the E, S and G dimensions. Their textual ESG scores explain a substantial share of the variance in leading commercial ratings, demonstrating that domain‑tuned NLP can markedly narrow the long‑standing "rating gap" in ESG measurement.
These two works do not address recent large-scale LLMs, and thus focus on task-specific fine-tuning instead of in-context-learning (as investigated in this work).

Other, more recent approaches tackle SDG information extraction with LLMs: \citet{garigliotti-2024-sdg} 
investigate an LLM-based RAG pipeline, which -- similar to our approach -- addresses the task of detecting textual passages containing Sustainable Developmental Goal (SDG) targets.
\citet{thulke2024climate} introduce \emph{ClimateGPT}, a 7B–70B‑parameter, climate‑specialised LLM family pretrained on 4.2B domain tokens and instruction‑tuned with expert tasks, matching Llama‑2‑70B on bespoke climate benchmarks while reducing hallucinations through retrieval augmentation.
While both approaches are based on LLMs, they use manual prompt tuning and do not address in-context learning from user feedback, which is the focus of our work.

\section{Approach}
\label{approach}

Given a text passage from a sustainability report, we classify whether this passage contains a relevant emission goal. To achieve this, we  prompt an instruction fine-tuned LLM,  specifically OpenAI's GPT-4o mini. The prompt includes both a set of task-specific instructions and -- optionally -- a small number of few-shot examples, i.e. input/output pairs that demonstrate the desired behavior. For a passage containing a relevant emission goal, the LLM's answer is expected to be "True," while irrelevant passages should be classified "False." Within this framework, we investigate two methods 
to include expert knowledge in form of labeled examples: \emph{Few-Shot Example Selection} and \emph{Automatic Prompt Design}.

\subsection{Few-shot Example Selection}
\label{exampleselection}

Here, we add labeled few-shot examples to the prompt, which are selected to be semantically similar to the target passage. 
For example, 
the sentence \textit{\enquote{Our goal is to reduce paper waste in our administrative departments by 35\% by 2027}} might be misclassified as a relevant goal. However, injecting a similar example with the correct label such as \textit{\enquote{By 2028, we aim to reduce paper usage in our offices by 50\% through digitalization}} into the prompt arguably improves the chance of a correct result. 


We embed passages using \emph{Sentence Transformer}~\cite{reimers19}, specifically \emph{all-MiniLM-L6-v2}, which balances efficiency and performance in MTEB \cite{muennighoff23}. This model is fine-tuned on 1.17 billion sentence pairs \cite{reimers_sbert} using contrastive learning. We retrieve similar passages via cosine similarity in the embedding space.


Since class distribution affects the LLM’s decision, we select up to three examples per class, prioritizing the most similar ones to form five few-shot demonstrations.

\subsection{Automatic Prompt Design}

Given the training set of labeled passages, we apply an approach similar to
\emph{ProTeGi}~(\citet{pryzant23}): Starting from an initial prompt, the LLM iterates through the training set of labeled text passages in random order. Given a text passage $p$, the current prompt -- consisting of instructions $\mathcal{I}$ and optionally some static few-shot demonstrations -- is used to predict whether $p$ contains a relevant emission goal. If the model's prediction contradicts the ground truth, we prompt the LLM to analyze possible root causes of the error.
This step is inspired by the \emph{Chain-of-Thought} method proposed by \citet{wei22}.
We then feed the  LLM's rationale back to the LLM, prompting it to modify $\mathcal{I}$ to correct the error so that the desired label is generated in future predictions. 
We obtain a new prompt candidate, comprising of new instructions $\mathcal{I}'$ (and optionally the same few-shot examples).

The accuracy of this new prompt with instructions $\mathcal{I}'$ is assessed by computing its F1 score on the training set.
 Should this exceed the score 
 of the previous instructions $\mathcal{I}$  
 by at least a small margin $\epsilon$, the new prompt is accepted and the iteration is continued with $\mathcal{I}'$. The margin $\epsilon$ ensures  an observed improvement to be statistically significant, and also limits the instruction complexity
(we found most modifications made by the LLM to add new clauses and/or sentences to the instruction). 
In contrast to the work by \citet{pryzant23}, we do not employ beam search 
but a simple greedy search. This is to limit the computational cost associated with high numbers of evaluations.

\section{Experiments}

We compare the effectiveness of the above approaches for  example selection and automatic instruction design in improving emission goal detection with LLM prompting.  

\paragraph{Dataset}


We use the NetZeroFact-BIG Dataset \cite{wrzalik24}, which contains passages from 16 business reports labeled by analysts. A passage is relevant (true) if it yields a correct fact; otherwise, it is false.



The data has been split into a test set (on which we estimate performance metrics) and a training set (from which we draw few-shot examples in example selection, and which the automatic prompt design iterates over).
 To prevent data leakage between the splits due to duplicate statements within the same report, 
 we split along the reports: Four reports have been chosen whose 207 passages form the test set, while 562 passages from the remaining 12 reports form the training set. 36\%/48\% of labels in the training/test split are positive.



\begin{table*}[ht!]
    \centering
    \caption{Results of Automatic Prompt Design (\%).}
\label{tab:autoprompt}
\begin{small}
    \begin{tabular}{llcccc|cccc}
    \toprule
    \multicolumn{2}{c}{\textsc{Examples during...}}& \multicolumn{4}{c}{simple start instruction} & \multicolumn{4}{c}{expert start instruction} \\
     \cmidrule(lr){1-2} \cmidrule(lr){3-6} \cmidrule(lr){7-10}
    Tuning & Testing & Acc & Prec & Rec & F1 & Acc & Prec & Rec & F1 \\
    \midrule
    \multicolumn{2}{c}{{\it (no tuning, zero-shot)}} & 56.2 & 52.2 & 100.0 & 68.6 & 81.2 & 77.1 & 86.1 & 81.4 \\
    \midrule
    Zero-shot & Zero-shot & 88.1 & 84.2 & 92.5 & 88.2 & 86.9 & 83.3 & 90.8 & 86.9 \\
    Zero-shot & Static & 83.2 & 80.4 & 85.9 & 83.0 & 87.0 & 92.2 & 79.7 & 85.4 \\
    Zero-shot & Random & 82.3 & 75.7 & 93.9 & 83.6 & 84.9 & 83.1 & 85.9 & 84.3 \\
    Zero-shot & Similar & 83.1 & 80.3 & 85.7 & 82.9 & 84.9 & 87.0 & 80.4 & 83.6 \\
    \midrule
    Static & Zero-shot & 87.4 & 90.1 & 82.7 & 86.2 & 89.9 & 88.2 & 90.9 & 89.5 \\
    Static & Static & 89.9 & 88.7 & 90.5 & 89.6 & 89.1 & 93.5 & 83.0 & 87.9 \\
    Static & Random & 81.7 & 76.0 & 91.3 & 82.8 & 89.0 & 88.0 & 89.9 & 88.6 \\
    Static & Similar & 84.1 & 84.2 & 82.3 & 83.2 & 87.4 & 90.5 & 82.4 & 86.3 \\
    \bottomrule
    \end{tabular}
    \end{small}
\end{table*}

\paragraph{Setup and Technical Details}
We have run all experiments for two versions of the starting instructions $\mathcal{I}$\footnote{All instructions and examples can be found in the appendix.} : (1) A {\it simple} ad-hoc version, and (2) an {\it expert} version, which resulted from a manual process of iterative optimization and result inspection prior to the experiments presented in this paper. 
As few-shot examples, we either use none ("Zero-shot"), $5$ random ones from the training set ("Random"), a fixed set of $5$ examples, which were expert-selected to be particularly informative prior to our experiments ("Static"), or example selection as described above ("Similar").

We report well-known quality metrics, namely classification accuracy,  precision, recall and the F1 score.
Our experiments were run with \textit{GPT-4o mini}\footnote{The version used is \texttt{gpt-4o-mini-2024-07-18}} through the \textit{OpenAI API} \cite{openai} with 
Python's \textit{LangChain} framework. 

We set the margin to $\epsilon{=}0.01$, which corresponds approximately to two times the standard deviation observed in many of our experiments (despite greedy generation, we found OpenAI's output to be non-deterministic, which is why we repeated evaluation $7$ times and report the average). With this margin, we found the prompt tuning process to converge in less than one epoch.


\paragraph{Few-Shot Example Selection}
Table \ref{tab:fewshot} displays test results for simple (left) and expert (right) prompts with the different few-shot example selection strategies. Note that only few-shot examples (and not the prompt) are varied in this experiment. 
We see that adding few-shot examples consistently improves results for the simple prompt, with expert-selected examples ("Static") complementing this prompt best. 
For the expert prompt, however, only the similarity-based example selection yields an improvement by $\approx 5$\%. We hypothesize that this complements the -- already quite elaborate -- prompt best. Also, note that few-shot prompting improves precision rather than recall (likely because examples emphasize intricacies of the extraction task).



\paragraph{Automatic Prompt Design}
Table \ref{tab:autoprompt}
illustrates results for automatic prompt design. 
First, and most importantly, we observe strong improvements over the manual prompts in the zero-shot case (Rows 1+2), from 68.6\% to 88.2\% for the simple prompt and from 81.4\% to 86.9\% for the expert prompt (note that -- surprisingly -- starting the optimization process from the simple prompt works even better). In both cases,
automatic prompt design outperforms example selection. The Appendix gives a qualitative impression of the evolution of the prompt, outlining which aspects were added in the optimization process, and that -- indeed -- the resulting  prompt reflects on some key intricacies of the extraction task.

Few-shot examples in this setting deteriorates results consistently when no examples were used in prompt tuning (Row 2 vs. Rows 3-5).  It seems that these examples confuse the model with its highly specific instructions. 
When using static examples in training\footnote{For efficiency reasons, we stick with static examples, since fixed prompts during prompt tuning allow for batching.}, we observe improvements in some cases but not with example selection (last row).

\section{Conclusion}
Overall, our results suggest that -- particularly with ad-hoc prompts, in which little explicit knowledge of the extraction task is encoded  -- automatic prompt design is  more effective compared to example selection.
A surprising finding is that, when applying automatic prompt design, we found few-shot samples, particularly when drawn with different strategies compared to prompt tuning, to be harmful in some cases. 

Future research on the issue may include experiments with other LLMs (particularly open-source or open-weight ones), the extension to other sustainability-oriented information extraction tasks (such as reported de-facto emissions, which often come in tables), and methods for interactive prompt-codesign by expert and LLM.

\section*{Limitations}
One key limitation of our study is that we only focus on OpenAI's o4-mini model, such that -- also due to the rapid advancement of large language models (LLMs) -- our findings merely represent a snapshot at the time of our experiments. While we conducted initial evaluations with more recent LLaMA-3 models~\cite{grattafiori2024llama3herdmodels} and observed similar performance trends as for the OpenAI-based experiments in this paper, findings may differ for newer models with enhanced reasoning capabilities such as DeepSeek-R1~\cite{deepseekai2025deepseekr1incentivizingreasoningcapability}. Specifically, these latest models have been claimed to offer improved performance on tasks requiring long reasoning chains. It should be, however,  that our specific task (emission goal detection) relies more on the precise assessment of edge cases rather than extended reasoning, suggesting that our core findings remain relevant despite these developments. Future work should systematically evaluate newer models to assess their potential impact on this task.

Another limitation of our study is the focus on company reports as the only -- and inherently non-objective -- source of information. While our extraction methods focus on identifying verifiable facts, the lack of external validation poses a risk of bias in the results. Here, a valuable direction for future research would be to cross-match extracted facts with independent sources, such as social media discussions, reports from non-governmental organizations (NGOs), or investigative journalism. This could provide a more comprehensive and balanced assessment of corporate emissions goals.

\section{Acknowledgements}
This work was supported by the German Federal Ministry of Education and Research (BMBF), Project "NetZeroNet" (01IS22050).

\bibliography{main}
\bibliographystyle{acl_natbib}

\appendix

\section*{Appendix}
\section{Prompts}
\lstset{
  basicstyle=\linespread{0.3}\footnotesize\ttfamily, 
  breaklines=true,
}

\subsection{Simple Instruction}
\label{simpleprompt}

\begin{lstlisting}
System: Determine if the text describes a commitment to reducing carbon emissions or achieving net zero; return "True" if it does, otherwise return "False".
Human: <example input>
AI: <example output>
...
Human: <input passage>
AI: <output prediction>
\end{lstlisting}

\subsection{Expert Instruction}
\label{expertprompt}

\begin{lstlisting}
System: You are an information extraction tool for climate goals that classifies whether a given text contains a statement about the commitment to a goal regarding carbon emissions. I will present you with passages from asset managers' reports. You will determine whether the given text contains a commitment to either a specific relative reduction in carbon emissions or to achieving net zero or carbon neutrality. Ignore any vague statements; a target is only a target if it states by when the target is to be achieved. For relative emission reductions, a specific percentage reduction must be stated. Ignore goals of third parties. Your answer is 'True' if the statement contains such a climate target of the asset manager and 'False' if it does not.
Human: <example input>
AI: <example output>
...
Human: <input passage>
AI: <output prediction>
\end{lstlisting}

\subsection{Static Few-Shot Examples}
\label{manualfewshot}

\begin{lstlisting}
System: <Instruction>
Human: A standout feature of the GreenTech Solutions Factory is its pledge to function entirely on renewable energy sources, aiming for net-zero emissions across its operations.
AI: False
Human: We are participating in the UN-backed Net-Zero Asset Owner Alliance (AOA) where a large number of the worlds biggest investors commit themselves to being net carbon neutral in their investments by no later than 2050 and to continually make five-year sub-targets for CO2 footprints.
AI: True
Human: Net zero emissions means achieving a balance between greenhouse gas (GHG) emissions produced and the amount removed from the atmosphere, consistent with limiting global warming to 1.5C and neutralising the impact of any residual emissions by permanently removing an equivalent amount of carbon dioxide (CO2). For BTPS this will mean reducing the portfolio's emissions through changing investments and investing in technologies which reduce emissions.
AI: False
Human: We have also emphasised our green ambitions by announcing that, from 2019-2025, we will reduce the carbon footprints of our investments by 29 per cent.
AI: True
Human: In our climate targets published in 2019, we are committed to exiting from investments in thermal coal by 2025. We are also committed to excluding oil exploration from our investments by 2030.
AI: False
Human: <input passage>
AI: <output prediction>
\end{lstlisting}

\subsection{Automatic Prompt Tuning: Prompt for Instruction Modification}

\begin{lstlisting}
System: <instruction>
Human: <example input>
AI: <example output>
...
Human: <input passage>
AI: <output prediction>
Human: Your prediction is wrong, we expect the answer to be "<target label>". Analyze the error in your previous prediction and identify which part of the instruction may have led to the incorrect result.
AI: <output reasoning>
Human: Modify the instruction to improve understanding of the expected behavior for the next prediction. Ensure the modification corrects the mistake without reversing the labels. Provide only the modified instruction in your response.
AI: <output instruction>

\end{lstlisting}

\subsection{Interaction Example from Automatic Prompt Engineering}

The following is a dialogue from the Automatic Prompt Engineering process, starting with the simple instruction and not including the few-shot examples during optimization.

\begin{lstlisting}
System: Determine if the text describes a commitment to reducing carbon emissions, achieving net zero, or setting specific emission reduction targets; return "True" if it does, otherwise return "False".
Human: 2 Guide for Identifying Sustainable Financing. 3 Identified Staff is made up of directors, senior managers or employees whose professional activities have a significant impact on the risk profile of an entity. An environmental and climate strategy that aims to contribute to the sustainable tran- sition, addressing the challenge of accelerating the transition to a carbon neutral economy, taking into account the natural capital.
AI: True
Human: Your prediction is wrong, we expect the answer to be "False". Analyze the error in your previous prediction and identify which part of the instruction may have led to the incorrect result.
AI: Upon reevaluating the text, it does not explicitly mention a commitment to reducing carbon emissions, achieving net zero, or setting specific emission reduction targets. Instead, it discusses an environmental and climate strategy aimed at contributing to a sustainable transition towards a carbon-neutral economy, which is more about the broader goal rather than specific commitments or targets.\n\nThe error in my previous prediction stemmed from interpreting the phrase "contribute to the sustainable transition" as a commitment, when it does not specify concrete actions or targets related to emission reductions. The instruction to identify specific commitments or targets was not adequately applied in this case.\n\nTherefore, the correct answer should indeed be "False." Thank you for your feedback.'
Human: Modify the instruction to improve understanding of the expected behavior for the next prediction. Ensure the modification corrects the mistake without reversing the labels. Provide only the modified instruction in your response.
AI: Determine if the text explicitly describes a commitment to reducing carbon emissions, achieving net zero, or setting specific, measurable emission reduction targets. Return "True" if it does, otherwise return "False." Focus on clear statements of intent or quantifiable goals rather than general strategies or aspirations.

\end{lstlisting}

\section{Automatic Prompt Tuning: Instruction Evolution}
\label{sec:instruct_evo}
\subsection{Initial Prompt: Simple Instruction - Zero-Shot}

\small \textbf{Initial Instruction:}
\textit{Determine if the text describes a commitment to reducing carbon emissions or achieving net zero; return "True" if it does, otherwise return "False".}
~\\

\small \noindent\textbf{Summarized Instruction Evolution:}
\begin{enumerate}
    \item Emphasizes \emph{specific} emission reduction targets.
    \item Focus on explicit statements of intent or measurable targets.
    \item Excludes vague or general strategies.
    \item Adds requirement for numerical values or timelines in commitments.
    \item Stresses actionable targets with clear timelines.
    \item Focuses on specific commitments, not just data or ambitions.
    \item Clarifies that numerical values alone are insufficient.
    \item Emphasizes intent to achieve targets within a timeframe.
    \item Notes need for long-term commitment or endpoint.
    \item Recognizes specific phrasing like "reduce to zero by 2050."
\end{enumerate}

\begin{small}
\noindent \textbf{Tuned Instruction:}
\textit{Determine if the text explicitly describes a commitment to reducing carbon emissions, achieving net zero, or setting specific, measurable emission reduction targets with clear numerical values or timelines. Return "True" if it does, otherwise return "False." Focus on clear, quantifiable statements of intent or goals, ensuring that any mentioned targets are specific, actionable, and include a defined timeline or endpoint. Emphasize that a commitment must include a direct statement of intent to achieve a specific target within a specified timeframe, rather than general goals or investments in technology without clear emission reduction outcomes. Note that numerical data or reduction ambitions alone, without a clear long-term commitment or endpoint, do not qualify as a commitment. Additionally, ensure to recognize that phrases indicating intent to achieve specific targets by a certain year (e.g., "reduce to zero by 2050") are valid commitments.}
\end{small}

\subsection{Initial Prompt: Manually Tuned Instruction - Zero-Shot}

\small \textbf{Initial Instruction:} 
\textit{You are an information extraction tool for climate goals that classifies whether a given text contains a statement about the commitment to a goal regarding carbon emissions. I will present you with passages from asset managers' reports. You will determine whether the given text contains a commitment to either a specific relative reduction in carbon emissions or to achieving net zero or carbon neutrality. Ignore any vague statements; a target is only a target if it states by when the target is to be achieved. For relative emission reductions, a specific percentage reduction must be stated. Ignore goals of third parties. Your answer is 'True' if the statement contains such a climate target of the asset manager and 'False' if it does not.
}
~\\

\small \noindent\textbf{Summarized Instruction Evolution:}
\begin{enumerate}
    \item Adds emphasis on the need for both a clearly stated percentage and a target year for relative reductions, and a target year for net zero/carbon neutrality. 
    \item Stresses the validity of the target only if it explicitly states the target year for both relative reductions and net zero/carbon neutrality. Includes checks to ensure the presence of a target year for all goals.
    \item Refines further by requiring that net zero or carbon neutrality commitments specify a particular year (e.g., 2030), not just a vague timeframe. Reinforced the need for specific target years and percentages for all commitments.
\end{enumerate}

\small \noindent \textbf{Tuned Instruction:}
\textit{You are an information extraction tool for climate goals that classifies whether a given text contains a statement about the commitment to a goal regarding carbon emissions. I will present you with passages from asset managers' reports. You will determine whether the given text contains a commitment to either a specific relative reduction in carbon emissions (with a clearly stated percentage and a target year) or to achieving net zero or carbon neutrality (with a specified target year). A target is only valid if it explicitly states by when the target is to be achieved, including a specific year (e.g., 2030) for net zero or carbon neutrality commitments. For relative emission reductions, a specific percentage reduction must be stated along with a target year. Ignore goals of third parties and any vague statements. Your answer is 'True' if the statement contains such a climate target of the asset manager and 'False' if it does not. Ensure to check for the presence of a clearly specified target year for both relative reductions and net zero or carbon neutrality commitments.}

\end{document}